\documentclass[11pt,a4paper]{article}
\usepackage[nohyperref]{emnlp2020}
\usepackage{times}
\usepackage{latexsym}
\usepackage{graphicx}

\usepackage{helvet}
\usepackage{courier}
\usepackage{amsmath}
\usepackage{url}
\usepackage{multirow}
\usepackage{booktabs}
\usepackage{enumitem}
\usepackage{amssymb}
\usepackage{marvosym}
\usepackage{color}
\usepackage{caption}
\usepackage{titlesec}
\usepackage{floatrow}
\newfloatcommand{capbtabbox}{table}[][\FBwidth]

\usepackage{microtype}

\aclfinalcopy 

\title{Multi-modal Summarization for Video-containing Documents}

\author{Xiyan Fu\textsuperscript{\rm 1}, Jun Wang\textsuperscript{\rm 2}, Zhenglu Yang\textsuperscript{\rm 1}\\
 \textsuperscript{\rm 1}College of Computer Science, Nankai University, China\\ 
\textsuperscript{\rm 2}Ludong University, China \\
\{fuxiyan,junwang\}@mail.nankai.edu.cn, \{yangzl\}@nankai.edu.cn}

\date{}

\begin{document}
\maketitle
\graphicspath{{graph/}}
\begin{abstract}

Summarization of multimedia data becomes increasingly significant as it is the basis for many real-world applications, such as question answering, Web search, and so forth. Most existing multi-modal summarization works however have used visual complementary features extracted from images rather than videos, thereby losing abundant information. Hence, we propose a novel multi-modal summarization task to summarize from a document and its associated video. In this work, we also build a baseline general model with effective strategies, \textit{i.e.,} bi-hop attention and improved late fusion mechanisms to bridge the gap between different modalities, and a bi-stream summarization strategy to employ text and video summarization simultaneously. Comprehensive experiments show that the proposed model is beneficial for multi-modal summarization and superior to existing methods. Moreover, we collect a novel dataset and it provides a new resource for future study that results from documents and videos. \footnote{If you are interested in our dataset, please contact authors.}

\end{abstract}

\section{Introduction}
Multi-modal summarization is conducted to refine salient information from different modalities (including text, image, audio, and video) and to represent key information through one or more modalities \citep{evangelopoulos2013multimodal, li2017multi-modal}. Given the rapid increase of multimedia data dissemination over the Internet, this task has been widely exploited in recent years.

Early works on multi-modal summarization has dealt with sports video summarization \citep{tjondronegoro2011multi}, meeting recordings summarization \citep{erol2003multimodal, li-etal-2019-keep}, or multimedia micro-blog summarization \citep{bian2013multimedia}. Most of these approaches concentrate on summarizing data that include synchronous information among different modalities. However, due to the lack of accurate alignment, summarizing the large volume of multi-modal data of the same topic is impractical. To address this issue, neural networks based methods have been introduced that search for the corresponding relations from different information resources. For example, \citet{li2017multi-modal} learn the joint representations of texts and images to identify the text that is relevant to an image. \citet{zhu2018msmo} and \citet{chen2018abstractive} propose an attention model to summarize a document and its accompanying images.

\begin{figure}[tp]
\centering\includegraphics[scale=0.5,trim=3 0 0 0]{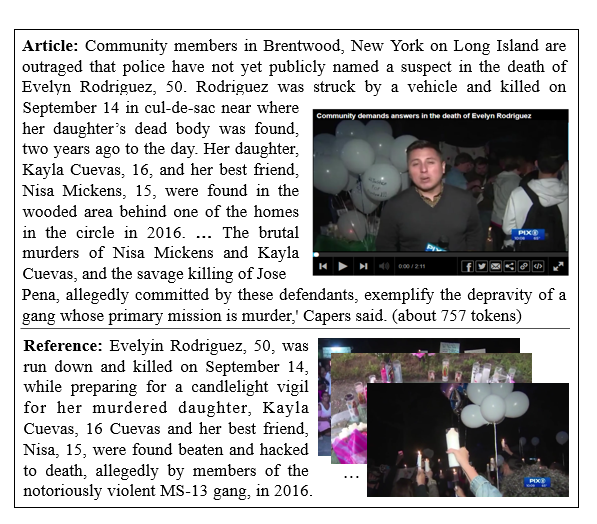}
\vspace{-10mm}
\caption{Illustration of the proposed multi-modal summarization task, which generates summaries from a document and its related video.}
\label{fig_example}
\end{figure} 

Although the aforementioned models have shown reasonable qualitative results, they still suffer from the following drawbacks: 1) Most existing applications extract visual information from the accompanying images, but they ignore related videos. We contend that videos contain abundant contents and have temporal characteristics where events are represented chronologically, which are crucial for text summarization. To the best of our knowledge, the only work \citep{li2017multi-modal} that considers video information, however, neglects temporal progression and simply treating videos as a bag of images. The authors use video frames for sentence matching in outputs rather than combining visual information with text processing.
2) Although attention mechanism~\citep{bahdanau2014neural} and early fusion~\citep{snoek2005early} are used extensively, it adversely
introduces noise as it is unsuitable for multi-modal data without alignment, which is characterized by a large gap that requires intensive communication.
3) Various multi-modal summarization works have focused on a single task, such as text or video summarization with added information from other modalities \citep{zhu2018msmo,wei2018video}. We observe that both summarization tasks share the same target of refining original long materials, and as such they can be performed jointly due to common characteristics.

In this work, we propose a novel multi-modal summarization task which is depicted in Fig.~\ref{fig_example}. To remove the noise among different modalities and effectively integrate complementary information, we introduce a bi-hop attention mechanism to align features and induce an improved late fusion method for feature fusion. Moreover, we apply a bi-stream summarization strategy for training by sharing the ability to refine significant information from long materials in text and video summarization. Given the lack of relevant datasets for experiments, we create a novel multi-modal dataset from the Daily Mail\footnote{\url{https://www.dailymail.co.uk/home/index.html}} and CNN \footnote{\url{https://www.cnn.com/}} websites, collecting articles and their corresponding summaries, videos, images.


The main contributions are as follows:
\begin{itemize}
\vspace{-3mm}
\item We introduce a novel task that automatically generates a textual summary with significant images from the multi-modal data associated with an article and its corresponding video.
\vspace{-3mm}
\item We propose the bi-hop attention and improved late fusion mechanism to refine information from multi-modal data. Besides, we introduce a bi-stream summarization strategy that simultaneously summarizes articles and videos.
\vspace{-3mm}
\item We prepare a content-rich multi-modal dataset. Comprehensive experiments demonstrate that complementary information from multiple modalities is beneficial, and our general baseline model can exploit them more effectively than the existing approaches.

\end{itemize}

\section{Related Work}

\textbf{Text summarization} selects salient information from long documents to form a concise summary. Existing approaches can be roughly categorized into two groups. 
\emph{Extractive-based methods}, which extract key sentences and objects without modification, guarantee basic-level grammaticality and accuracy. 
For example, \citet{jadhav2018extractive} modeled the interaction between keywords and salient sentences by using a new two-level pointer network. The proposal of~\citep{narayan2018ranking, wu2018learning} extracted summary via deep reinforced learning.
\emph{Abstractive-based methods}, which paraphrase the significant contents after comprehending the original document, construct sophisticated summaries with newly-generated words and coherent expressions.
~\citet{tan2017abstractive} proposed a graph-based attention mechanism in a sequence-to-sequence framework. 
The work of ~\citet{cao2018retrieve} retrieved proper existing summaries as candidate soft templates and extended the framework to jointly perform template reranking and summary generation. \citet{see2017get} presented a pointer-generator network which can both copy words from articles and generate novel words. 
Recently, the sheer volume of fine-tuning approaches \citep{liu-lapata-2019-text, zhang-etal-2019-hibert, dong2019unified} boosted the quality of the generated summaries based on the pre-trained language models which have advanced a wide range of NLP tasks. ~\par


\textbf{Video summarization} is conducted to facilitate large-scale video browsing by producing concise summaries. Early works in video summarization have focused on videos of certain genres. They generated summaries by leveraging genre-specific information, i.e., salient objects of sports in ~\citep{ekin2003automatic} and significant regions of egocentric videos in~\citep{lu2013story}. Furthermore, summarizing videos with auxiliary resources, such as web images and videos, has attracted considerable attention. \citet{song2015tvsum} developed a co-archetypal analysis technique that learns canonical visual concepts shared between a video and images. Given that the above mentioned non-deep summarization methods are time-consuming, ~\citet{zhou2018deep} modeled video summarization via a deep RNN to capture long-term dependencies in video frames and proposed a reinforcement learning-based framework to train the network end-to-end. Recently, a novel query-focused video summarization task was introduced in \citep{xiao2020convolutional}.

\textbf{Multi-modal summarization} generates a condensed multimedia summary from multi-modal original inputs, such as text, images, videos, and etc. \citet{UzZaman2011multimodal} introduced an idea of illustrating complex sentences as multimodal summaries that combine pictures, structure and simplified compressed text. \citet{libovicky2018multimodal, palaskar-etal-2019-multimodal} studied abstractive text summarization for open-domain videos. Besides, \citet{li2017multi-modal} developed an extractive multi-modal summarization method that automatically generates a textual summary based on a set of documents, images, audios and videos related to a specific topic. \citet{zhu2018msmo, zhu3multimodal} combined image selection and output to alleviate the modality bias. \citet{chen2018abstractive} employed hierarchical encoder-decoder model to align sentences and images. Recently, aspect-aware model was introduced in \citep{li-aspect} to incorporate visual information for producing summary for e-commerce products.

\section{Model}
\begin{figure}[t]
\centering\includegraphics[scale=0.33,trim=3 0 0 0]{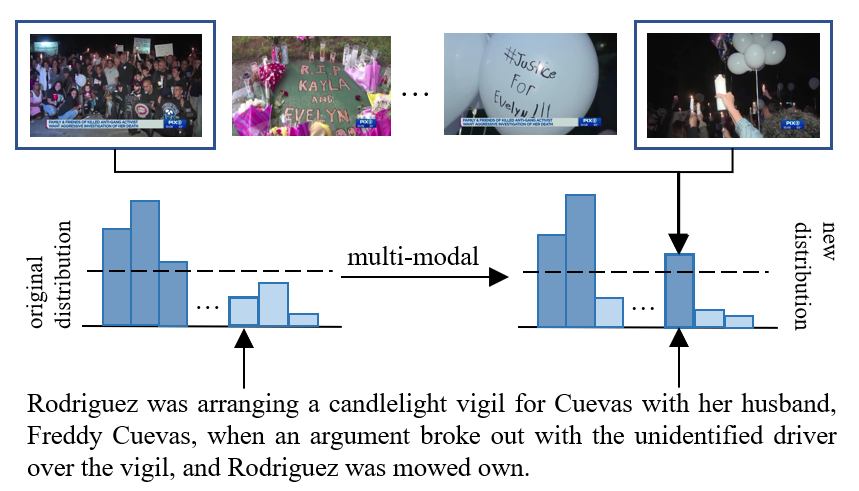}
\caption{$\rm M^{2}$SM utilizes multi-modal information for sentence extraction in summarization. The probability of the mismatched sentence calculated solely based on text is improved with the assistance of video.}
\vspace{-3mm}
\label{fig_effect}
\end{figure}

\begin{figure*}[t]
\centering\includegraphics[scale=0.65,trim=3 0 0 0]{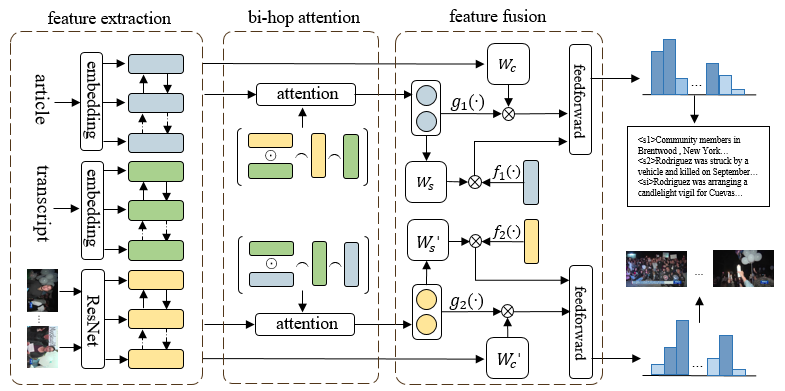}
\setlength{\belowcaptionskip}{-95pt}
\vspace{-4mm}
\caption{Overview of the multi-modal summarization model. $\odot$ and $\otimes$ respectively represent the Hadamard product and outer product, and $^\frown$ stands for vector concatenation. $f_{1}, f_{2}$ and $g_{1}, g_{2}$ are the special versions of $f(\cdot), g(\cdot)$ mentioned in ``Feature Fusion" section with different parameters.\vspace{-5mm}}
\label{fig_model}

\end{figure*}

We introduce a novel multi-modal summarization model ($\rm M^{2}$SM) to automatically generate multi-modal summary from an article and its corresponding video. Fig.~\ref{fig_effect} illustrates the idea.
Our proposed model consists of four modules: feature extraction, alignment, fusion, and bi-stream summarization, which are presented in Fig.~\ref{fig_model}.~\par


\subsection{Feature Extraction}
\textbf{Text Feature} We utilize a hierarchical framework based on word and sentence levels to read the tokens of an article and induce a representation of each sentence. The bi-directional long short term memory (BiLSTM) which constitutes a forward and a backward LSTM is employed as recurrent unit.
The first layer of the encoder, which extracts fine-grained information of a sentence, runs at the word level. The hidden state of the $j^{th}$ word in the $i^{th}$ sentence is represented by the BiLSTM, i.e., $h_{i}^{j}$=$[\overrightarrow{h_{i}^{j}},\overleftarrow{h_{i}^{j}}]$. Moreover, the encoder consists of a second layer conducted at the sentence level, which accepts the average-pooled, concatenated hidden states from the word level encoder as inputs. The hidden state of the $i^{th}$ sentence is defined as
\vspace{-3mm}
\begin{equation}
s_{i}=[\phi(ap_{i},\stackrel{\rightarrow}{s}_{i-1}),\phi(ap_{i},\stackrel{\leftarrow}{s}_{i+1})],
ap_{i}=\sum_{j=1}^{N_{i}}h_{i}^{j},
\vspace{-2mm}
\end{equation}
where $N_{i}$ is the number of words in the $i^{th}$ sentence and $\phi(\cdot)$ represents LSTM.

\noindent \textbf{Video Feature} Videos that accompany articles often capture news highlights and usually provide abundant complementary information from the perspective that is different from the article itself.
We choose ResNet-50~\citep{he2016deep} to extract frame features for its excellent performance.
Furthermore, we train a BiLSTM to model the sequential pattern in video frames, where the distinctions between videos and bag of images are exhibited. The image feature of the $i^{th}$ frame is a 2048 dimensional vector $v_{i}$, which is obtained using $v_{i}$=ResNet$(frame_{i})$.
Correspondingly, the representation of a frame is defined as
\vspace{-2mm}
\begin{equation}
m_{i}=[\varphi(v_{i},\stackrel{\rightarrow}{m}_{i-1}),\varphi(v_{i},\stackrel{\leftarrow}{m}_{i+1})],
\vspace{-2mm}
\end{equation}
where $\varphi(\cdot)$ represents BiLSTM. 

Recently, inspired by the success of the language model 
BERT \citep{bert2019}, which pre-trains deep bidirectional representations 
on unlabeled texts, 
a few visual-linguistic joint models 
have been proposed, such as VideoBERT~\citep{sun2019videobert}, ViLBert~\citep{lu2019vilbert}, and VLP~\citep{zhou2019unified}. 
These models can 
induce superior features 
for summarization.
Our model $\rm M^{2}$SM is devised from a new perspective to extract fine-grained features, which are orthogonal to the aforementioned BERT-based ones in a complementary manner.


\subsection{Feature Alignment}
Existing multi-modal models with high alignment cannot be used for capturing the correct common features among different modalities 
due to the asynchronism between modalities. Hence, we introduce two multi-modal attention mechanisms which focus on diverse parts of video frames in accordance to a current article sentence in order to search for aligned information. ~\par
\noindent\textbf{Single-step Attention} We fulfill the above-stated goal by adopting the attention mechanism which was introduced in~\cite{bahdanau2014neural}. Specifically:
\vspace{-2mm}
\begin{equation}
e_{i}^{j} = V^{T}\tanh(W_{s}s_{i}+W_{m}m_{j}+b_{attn}), 
\vspace{-2mm}
\end{equation}
where $e_{i}^{j}$ is the attention value between the $i^{th}$ sentence and $j^{th}$ image, and $V$, $W_{s}$, $W_{m}$, and $b_{attn}$ are learnable parameters. Although lots of multi-modal summarization models \citep{chen2018abstractive,zhu2018msmo,li2018multi-modal_sent} follow this mechanism, we argue that it is unsuitable for a multi-modal task because 
one modality might dominate the summary resulting in substantial loss of information from other modalities. Therefore, we generalize over ``bilinear" attention \citep{kim2017hadamard} and propose improvements such as: 

\begin{itemize}[leftmargin=*]
\vspace{-2mm}
\item Single feature projection~\par
Considering that features of different modalities are independent from each other, the projection can be separately calculated such that neither component dominates another. The feature mapping of each modality can be modified as follows:
\vspace{-2mm}
\begin{align}
\label{eq.4}&q_{j} = \tanh(W_{m}m_{j}+b_{m}),\\
\label{eq.5} &r_{i} = \tanh(W_{s}s_{i}+b_{s}),
\end{align}
\noindent where $W_{m},W_{s},b_{m}$, and $b_{s}$ are parameters.
\vspace{-3mm}
\item Dual feature interaction~\par
For close communication of different modalities, we further propose a dual interaction feature calculated using Hadamard product. The concatenation feature \textit{cf} 
represents the feature of each modality,
while dual interaction feature \textit{dif} avoids insufficient interaction among modalities. The attention value is formulated as
\vspace{-2mm}
\begin{equation}
\vspace{-3mm}
e_{i}^{j} = V^{T}(\underbrace{q_{j}\odot r_{i}}_{\text{dif}} +\underbrace{q_{j}+r_{i}}_{\text{cf}}).
\end{equation}
\end{itemize}
\noindent After calculating the softmax of the attention value to obtain weight $a_{i}^{j}$ of each image state $m_{j}$, and computing the weighted sum of these states, context value can be represented as follows: 
\vspace{-3mm}
\begin{equation}
\vspace{-3mm}
c_{i} = \sum_{j=1}^{NM}a_{i}^{j}m_{j}, \quad  a_{i}^{j} = softmax(e_{i}^{j}),
\end{equation}

\noindent where \textit{NM} is the number of filtered frames.

\noindent\textbf{Bi-hop Attention}
Given that the transcript extracted from a video shares the same modality with an article and accurately aligns with a video, we induce it as a bridge to deepen the relationship between two modalities. We introduce bi-hop attention to produce a context by simultaneously combining text sentence with a transcript and video frames. Since a transcript is similar to an article text, we use the BiLSTM to extract its features $t_{1}, t_{2}, ..., t_{NT}$, where \textit{NT} is the number of words in transcripts. Moreover, the context vector of transcript $ct2m$ is obtained by replacing $s_{i}$ with $t_{i}$ in Eq.~\eqref{eq.5} for a new attention value $b_{i}^{k}$, and the original context vector of article $ca2m$ is improved as $ca2t$ by replacing $m_{j}$ with $ct2m_{j}$ in Eq.~\eqref{eq.4} for $d_{k}^{j}$:
\vspace{-5mm}
\begin{equation}
ca2m_{i} = \sum_{k=1}^{NT}b_{i}^{k}ct2m_{k},\qquad \\
\; ct2m_{k} = \sum_{j=1}^{NM}d_{k}^{j}m_{j}.
\end{equation}

\noindent 
\noindent Similarly, bi-hop attention can be reversed to obtain an article context for video summarization.

\subsection{Feature Fusion}
In terms of combining complementary information from multiple modalities, we develop a method that not only smoothly suppresses the effect of modality in unfavorable situations, but also captures synergies between the modalities that share reliable complementary information. Methodologically, we use two common strategies for taking cross-modal correlations: \textbf{Early fusion}, which concatenates various features directly, has been explored in multi-modal tasks extensively \citep{zhu2018msmo, chen2018abstractive}. The fusion feature of \textit{$i^{th}$} sentence of text summarization can be represented as $p_{ts}^{i}=[s_{i},ca2m_{i}]$. \textbf{Tensor fusion} \citep{zadeh-etal-2017-tensor} disentangles unimodal and bimodal dynamics by modeling each of them explicitly for intra-modality and inter-modality dynamics. $p_{ts}^{i}=[s_{i}1]\otimes [ca2m_{i}1]$ is denoted as fusion feature, where $\otimes$ indicates the outer product. Further, the label of each sentence can be predicted by $p_{ts}^{i}$. ~\par
Both the above-mentioned strategies are based on a strong assumption that each modality is accurately aligned, yet multi-modal summarization always contains asynchronous information.
Hence, we consider that \textbf{late fusion}, which uses unimodal decision values and fuses them with a fusion mechanism, suits multi-modal summarization well. $F(f(s_{i}), g(ca2m_{i}))$ is the fusion process where $f(\cdot),g(\cdot)$ is a conventional feed forward network, and $F(\cdot)$ can be a function, such as averaging, voting or other learned models. We use a feedforward network in this work. Given that each modality is not useful all the time, e.g., frames of accompanied interview video may contribute only to a small extent to visual features, we restrain the noise from a certain modality by following the ideas in \citep{liu2018learn} and induce \textbf{late+ fusion}. Its fusion process is improved as $F(W_{s}f(s_{i}), W_{c}g(ca2m_{i}))$ and $W_{s}, W_{c}$ are calculated as follows:
\vspace{-2mm}
\begin{equation}
W_{s}=[1-g(ca2m_{i})]^\beta, W_{c}=[1-f(s_{i})]^\beta,
\vspace{-2mm}
\end{equation}
\noindent where $\beta$ is a smoothing coefficient determining the penalty intensity.
For example, if a frame comes from an interview video and its visual features are irrelevant for classification, $g(ca2m_{i})$ becomes small which bases the prediction mainly on article.

\subsection{Bi-stream Summarization Training}
Given that text and video summarization aim to extract salient information from the original redundant content, we propose bi-stream summarization training strategy. It indicates that they are learned jointly and simultaneously, which improve the generalization performance by holding similar objectives and sharing complementary information. ~\par 

We consider extractive-based text summarization as a sentence classification task, in which the binary label of each sentence is imperative. Given that most corpora only contain abstractive summaries written by humans, we construct the label of each sentence following the methods in \citep{nallapati2017summarunner}. Sentences are selected to maximize the ROUGE with respect to the gold summary by a greedy approach. Furthermore, we use cross entropy for training:
\vspace{-2mm}
\begin{equation}
\mathcal{L}_{ts}=-\frac{1}{NS} \sum_{n=1}^{NS}\left[y_{n} \log \hat{y}_{n}+\left(1-y_{n}\right) \log \left(1-\hat{y}_{n}\right)\right],
\vspace{-2mm}
\end{equation}
\noindent where $y_{n}$ and $\hat{y}_{n}$ represent the true and predicted labels, respectively.

Video summarization can also be considered as a classification task; we use unsupervised learning by RL methods proposed in~\citep{zhou2018deep}.
Its loss can be separated into diversity reward $R_{div}$ and representativeness reward $R_{rep}$. The former measures the dissimilarity among selected frames in the feature space, whereas the latter measures how well the generated summary can represent the original video. The calculations are as follows:
\vspace{-2mm}
\begin{equation}
\begin{aligned}
R_{\mathrm{div}}=\frac{1}{|\mathcal{M}|(|\mathcal{M}|-1)} \sum_{j \in \mathcal{M}} \sum_{j^{\prime} \in \mathcal{M} \atop j^{\prime} \neq j} d\left(m_{j}, m_{j^{\prime}}\right) \\
R_{\mathrm{rep}}=\exp \left(-\frac{1}{NM} \sum_{j=1}^{NM} \min _{j^{\prime} \in \mathcal{M}}\left\|m_{j}-m_{j^{\prime}}\right\|_{2}\right)
\end{aligned}
\vspace{-2mm}
\end{equation}

\noindent where $\mathcal{M}$ is the set of selected video frames, and $d(\cdot)$ is the dissimilarity function, which employed as one minus cosine similarity in this work.

For bi-stream summarization, we mix the loss from dual summarization tasks, whose contribution relies on their weight $\alpha_{ts}$ and $\alpha_{vs}$:
\vspace{-3mm}
\begin{equation}
\mathcal{L} = \alpha_{ts}\mathcal{L}_{ts} + \alpha_{vs}(R_{\mathrm{div}}+R_{\mathrm{rep}}).
\vspace{-2mm}
\end{equation}

\section{Experiments}
\subsection{Dataset and Evaluation Metrics}
There is no existing dataset that contains articles, corresponding videos and references for multi-modal summarization, we construct a new corpus called MM-AVS. To obtain high quality summaries, we collect data from Daily Mail and CNN websites same as in \citep{hermann2015teaching}. 
We also preserve the related titles, images and their captions for multi-modal research. 
Samples which miss any elements mentioned above or which video duration is less than 30 seconds are removed. 
The detailed statistics of the corpus are shown in Table \ref{tab_corpus_statistics}. Notice that the videos in CNN are much longer than that in Daily Mail, we collect less samples from the former and we mainly use them during testing. ~\par

ROUGE~\citep{lin2004rouge} with standard options is used for text summary evaluation. We apply the ROUGE-1 (R-1), ROUGE-2 (R-2), and ROUGE-L (R-L) F1-score to evaluate the overlap of uni-grams, bi-grams, and the longest common subsequences between the decoded summary and the reference. To evaluate the quality of the generated video summaries (images), we employ ResNet to construct image feature vectors and calculate the cosine image similarity (Cos) between image references and the extracted frames from videos.

\subsection{Baseline Methods}
\noindent \textbf{Multi-modal based:}

\noindent \textbf{VistaNet} \citep{vistanet}
prioritizes visual information as alignment to point out the important sentences of an article to detect the sentiment of a document.~\par 

\noindent \textbf{MM-ATG} \citep{zhu2018msmo}
is a multi-modal attention model generating text and selecting the relevant image from the article and alternative images.~\par

\noindent \textbf{Img+Trans} \citep{hori2019end} 
applies multi-modal video features including video frames, transcripts and dialog context for dialog generation.

\noindent \textbf{TFN} \citep{zadeh-etal-2017-tensor}
learns both intra and inter modalities by modeling a new multi-modal tensor fusion approach.~\par

\noindent \textbf{HNNattTI} \citep{chen2018abstractive} 
aligns the sentences and accompanying images by attention. ~\par

 
\noindent\textbf{Pure video summarization:}

\noindent \textbf{Random}
extracts the key video frames randomly. ~\par

\noindent \textbf{Uniform}
samples videos uniformly. ~\par

\noindent \textbf{VSUMM} \citep{de2011vsumm} extracts color features from video frames via k-means clustering.\par

\noindent \textbf{DR-DSN} \citep{zhou2018deep}
proposes a reinforcement learning framework equipped with a reward function for diverse and representative summaries. ~\par

\noindent \textbf{Pure text summarization:}

\noindent \textbf{lead3} picks the first three sentences as summary.

\noindent \textbf{SummaRuNNer} \citep{nallapati2017summarunner}
is a interpretable RNN-based sequence model, and it can be trained in both extractive and abstractive manners.~\par

\noindent \textbf{NN-SE} \citep{cheng2016neural} consists of a hierarchical document encoder and an attention-based extractor that can extract sentences or words.

\begin{table}[t]
\centering 
\resizebox{0.7\textwidth}{!}{
\begin{tabular}{llll} \hline
 &Daily Mail  &CNN  \\ \hline
Article Number &1970  &203 \\
Avg. Article Length &657.87  &951.07 \\
Avg. Article Sent. Num. &33.42  &29.84 \\
Avg. Summary Length &59.63  &29.73 \\
Avg. Video Duration &81.96  &368.19 \\
Avg. Transcript Length &83.74 &116.76 \\ \hline
\end{tabular}}
\caption{Corpus statistics. \vspace{-4mm}}
\vspace{-3mm}
\label{tab_corpus_statistics}
\end{table}

\subsection{Experimental Settings}
The dataset is split by 70\%, 10\%, 20\% for train, validation and test sets, respectively. The hidden dimension of Bi-LSTM is 64, beta in late+ fusion is 0.3, and the proportion of each modality in loss is 3.33. To remove the redundancy of videos, one of five consecutive frames is randomly selected. The last layer of ResNet-50 with 2048 dimension is used for image feature extraction. We perform training by Adagrad with learning rate 0.0001. We use early stop to avoid overfitting.

\begin{table*}
\resizebox{0.6\textwidth}{!}{
\begin{tabular}{lccccccc} \hline
\multirow{2}{*}{Model} & \multicolumn{4}{c}{Daily Mail} & \multicolumn{3}{c}{CNN} \\ \cmidrule(r){2-5}\cmidrule(r){6-8}
&R-1  &R-2  &R-L &Cos(\%)  &R-1 &R-2 &R-L \\ \hline
VistaNet  &18.62 &6.77 &13.65 &-  &9.31 &3.24 &6.33  \\ 
MM-ATG  &35.38  &14.79  &25.41 &69.17  &26.83 &8.11 &18.34  \\
Img+Trans  &39.28  &16.64  &28.53  &- &27.04 &8.29 &18.54  \\
TFN  &39.37  &16.38  &28.09 &-   &27.68 &8.69 &18.71    \\  
HNNattTI  &39.58  &16.71 &29.04 &68.76  &27.61 &8.74 &18.64  \\ \hline
$\rm M^{2}$SM &\textbf{41.73} &\textbf{18.59} &\textbf{31.68} &\textbf{69.22}  &\textbf{27.81}  &\textbf{8.87}  &\textbf{18.73}   \\ \hline
\end{tabular}}
\setlength{\abovecaptionskip}{1pt} 
\vspace{-3mm}
\caption{Comparisons of proposed $\rm M^{2}$SM model with the multi-modal baselines. 
All of the text encoder parts in multi-modal approaches are modified as hierarchical framework for fair comparison. \vspace{-4mm}}
\label{tab_main_results}
\end{table*}

\begin{table}[]
\centering
\resizebox{0.4\textwidth}{!}{
\begin{tabular}{lc} \hline
Model &Cos(\%)  \\ \hline
VSUMM  &68.74  \\
Random &67.69   \\ 
Uniform &68.79   \\
DR-DSN &68.69   \\ \hline
$\rm M^{2}$SM &\textbf{69.22}  \\ \hline
\end{tabular}}
\setlength{\abovecaptionskip}{1pt} 
\vspace{-3mm}
\caption{Comparisons of $\rm M^{2}$SM model with video summarization baselines. \vspace{-3mm}}
\label{tab_video_summary}
\end{table}

\begin{table}[t]
\centering
\resizebox{0.8\textwidth}{!}{
\begin{tabular}{lccc} \hline
Model &R-1  &R-2  &R-L \\ \hline
lead3  &41.07 &17.87 &30.90 \\
SummaRuNNer &41.12 &17.92 &30.94 \\
NN-SE &41.22 &18.15 &31.22 \\ \hline
$\rm M^{2}$SM &\textbf{41.73} &\textbf{18.59} &\textbf{31.68}  \\ \hline
\end{tabular}}
\setlength{\abovecaptionskip}{1pt} 
\setlength{\belowcaptionskip}{-40pt}
\vspace{-3mm}
\caption{Text summary comparisons of proposed $\rm M^{2}$SM model with the text summarization baseline. \vspace{-4mm}}
\label{tab_text_summary}
\end{table}

\section{Results}
\subsection{Quantitative analysis}
As shown in Table \ref{tab_main_results}, $\rm M^{2}$SM outperforms the best performing baseline for both text and video summaries on the Daily Mail dataset.
These improvements are achieved thanks to the bi-hop attention mechanism, improved late+ fusion, and bi-stream summarization strategy, which are all jointly incorporated in $\rm M^{2}$SM. Table \ref{tab_main_results} also shows that \textit{VistaNet} underperforms other models by a large margin. We consider that the learning strategy of \textit{VistaNet} wherein image information is considered prior to text information may bring noise.\par

In addition, Table~\ref{tab_main_results} shows that the results on CNN are inferior to those on Daily Mail. We consider this may be attributed to the longer length of original materials, especially of videos, as shown in Table ~\ref{tab_corpus_statistics}. Given that it is difficult to extract visual features from long videos, summarizing the articles in the CNN dataset is more challenging. Besides, we ignore image evaluation in CNN due to the lack of reference from the original website, and explore video summaries by comparing with pure video summarization in Section 5.3.

\subsection{Ablation Study}
To verify the effectiveness of each component of our model, we conduct ablation experiments. We construct a hierarchical framework which concentrates on word and sentence level to generate summaries as our baseline. Based on the hierarchical text component, three constituents of $\rm M^{2}$SM are added in turn: \textit{+video frames} indicates adding the frames of the related videos to provide complementary information, \textit{+transcripts} extracts transcripts of the videos to deepen the relationship between different modalities, and \textit{+bi-stream} uses the bi-stream summarization training strategy to learn the similarity shared by summarization tasks.

As shown in Table~\ref{tab_ablation}, each component of $\rm M^{2}$SM contributes to improving the performance by fusing video with text information. This validates our assumption that videos possess abundant complementary information to texts, and this information facilitates capturing the core ideas of materials comprehensively and inducing effective summaries.

\begin{table}[t]
\centering 
\resizebox{0.8\textwidth}{!}{
\begin{tabular}{lllll} \hline
 &R-1 &R-2 &R-L \\ \hline
text-only(baseline) &39.11 &16.42 &28.56 \\
+video frames &40.86 &17.48 &30.23 \\
+transcripts &41.26 &17.95 &30.98 \\
+bi-stream &41.73 &18.59 &31.68 \\ \hline
\end{tabular}}
\setlength{\abovecaptionskip}{1pt} 
\vspace{-3mm}
\caption{Ablation study
to evaluate the effects of different components of $\rm M^{2}$SM. \vspace{-4mm}}
\label{tab_ablation}
\end{table}

\subsection{Evaluation on Single Modalities}
To assess the gains of the proposed method coming from the multi-modal information, we compare it with the popular single-modal approaches on Daily Mail dataset, including the video only summarization and text only summarization. 

The video summary comparisons are shown in Table~\ref{tab_video_summary}. We collect the related images as references in an online manner, because manually labeling each video frame is labor-intensive and time-consuming.
As shown in Table~\ref{tab_video_summary}, $\rm M^{2}$SM performs better than the video only methods.
It can be attributed to its capability to derive comprehensive insights from multi-modal materials.

In terms of the text summary as shown in
Table~\ref{tab_text_summary}, $\rm M^{2}$SM is also competitive, although it does not achieve as much large margin as in the case of the multi-modal summary comparisons. We speculate this may be due to the noisy information. Using an effective information filter is a promising way which worth in-depth exploration in the future.

\begin{figure}[t]
\centering\includegraphics[scale=0.26,trim=0 0 0 0]{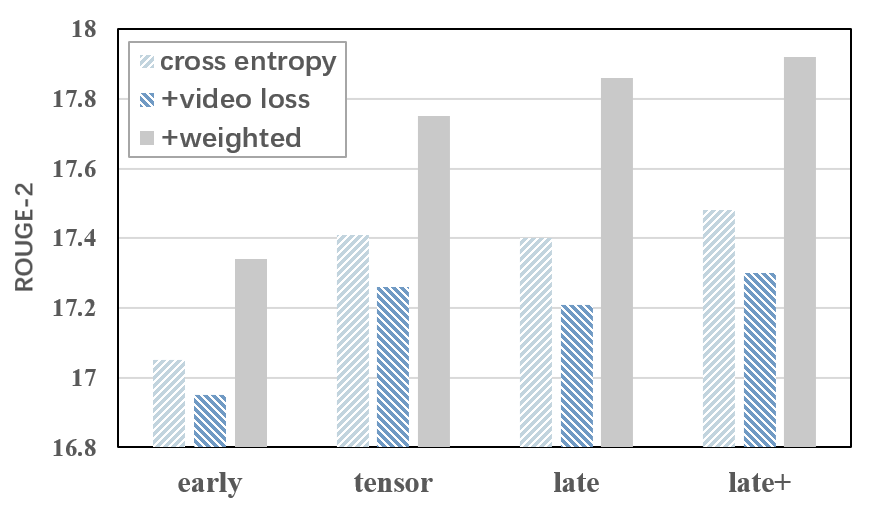}
\setlength{\abovecaptionskip}{-2pt} 
\vspace{-4mm}
\caption{Comparison of four feature fusion methods. Each way is conducted with three training strategies. \vspace{-4mm}}
\label{fig_fusion}
\end{figure}

\subsection{Sub-module Evaluation}
We evaluate the effectiveness of each module of the proposed model. Fig.~\ref{fig_fusion} illustrates the performance of four feature fusion approaches, which are early, tensor, late, and late+ fusion.
All these feature fusion methods are trained under three strategies: \textit{cross entropy} minimizing the binary cross entropy, \textit{+video-loss} adding video summarization loss for optimization, and \textit{+weighted} that gives weights to each task in consideration of their different contributions. As demonstrated in Fig.~\ref{fig_fusion}, early fusion, which is most commonly used in practice, is inferior to other fusion approaches. 
Due to the asynchrony of multi-modal feature and the various significance of each modality, simple feature concatenation in early fusion may introduce noise and lose its effectiveness for handling complex cases. 
Moreover, Fig.~\ref{fig_fusion} depicts that \textit{+weighted} achieves the best results. We consider that it successfully avoids unfavorable influence of irrelevant modalities.

\begin{table}[]
\centering 
\resizebox{0.8\textwidth}{!}{
\begin{tabular}{lllll} \hline
 &R-1 &R-2 &R-L \\ \hline
no-attention &40.13 &17.21 &29.85 \\
concat-product &39.59 &16.71 &29.04 \\ \hline
bilinear-attention &40.86 &17.48 &30.23 \\ \hline
\end{tabular}}
\setlength{\abovecaptionskip}{-1pt} 
\vspace{-3mm}
\caption{Comparison of three feature alignment strategies on the basis of early fusion. \vspace{-4mm}}
\label{tab_alignment}
\end{table}

\begin{figure}[t]
\setlength{\abovecaptionskip}{1pt} 
\centering\includegraphics[scale=0.28,trim=0 0 0 0]{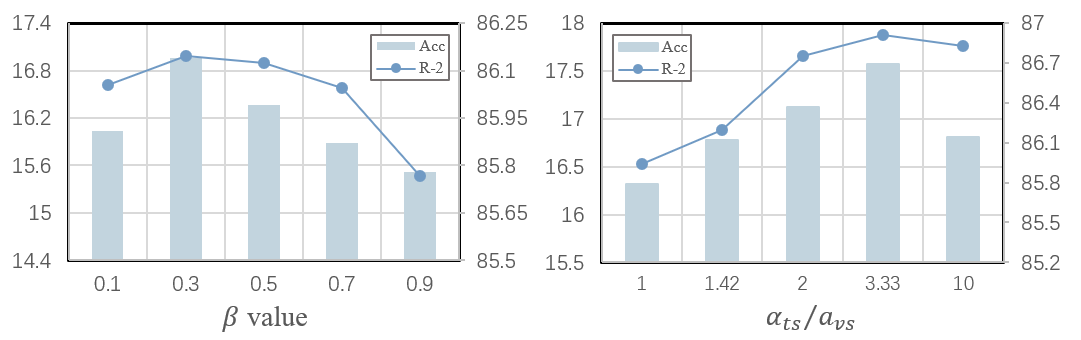}
\vspace{-8mm}
\caption{Evaluation of the smoothed value $\beta$ in late+ fusion and $\alpha_{ts}/\alpha_{vs}$ of each modality in loss. \vspace{-5mm}}
\label{fig_parameters}
\end{figure}

\begin{table}[t]
\centering
\resizebox{0.8\textwidth}{!}{
\begin{tabular}{lllll} \hline
\multicolumn{2}{l}{} &R-1 &R-2 &R-L  \\ \hline
\multirow{2}{*}{article} &CNN  &25.29  &5.70  &11.11  \\
                  &Daily Mail  &8.78 &2.15 &5.45  \\ \hline
\multirow{2}{*}{reference} &CNN  &9.67  &1.93  &6.23  \\
                  &Daily Mail  &6.02 &0.84 &4.18  \\ \hline
\end{tabular}}
\setlength{\abovecaptionskip}{1pt} 
\vspace{-3mm}
\caption{Words overlap statistics of video transcripts with articles and references. \vspace{-4mm}}
\label{tab_transcript}
\end{table}

\begin{table}[!t]
\centering 
\resizebox{0.6\textwidth}{!}{
\begin{tabular}{llll} \hline
 &Inform &Satis\\ \hline
article &3.65 &3.76 \\
video &2.73 &2.78\\
article+video &3.87 &4.30 \\ \hline
\end{tabular}}
\setlength{\abovecaptionskip}{1pt} 
\vspace{-3mm}
\caption{Manual Summary quality evaluation. \vspace{-4mm}} 
\label{tab_manual}
\end{table}

In addition, three attention mechanisms are tested: \textit{no attention} which uses the last state as video feature directly, \textit{concat-product} which uses the common mechanism introduced in \citep{bahdanau2014neural} and \textit{bilinear-attention} which we propose in this paper. Table ~\ref{tab_alignment} presents the results and indicates a stimulating phenomenon that concat-product attention mechanism performs poorly, even worse than models without attention. We speculate that conventional attention which lacks communication between modalities will mislead the model, as it focuses on irrelevant parts and brings noise.

\subsection{Balance between modalities}
A particular characteristic of multi-modal summarization is that the related data are complementary and thus different modalities contribute differently to summarization. Hence, we restrain the noises from irrelevant modalities through the improved late fusion and balance the loss function in different tasks. The left graph in Fig.~\ref{fig_parameters} illustrates the penalty of irrelevant information, which shows that $\beta=0.3$ yields the best results. The other graph depicts the proportion of each modality in a loss function, and illustrates that the text summarization gives approximately 70\% attention to the text modality. However, this graph also reveals that nearly 30\% of useful information is searched and complemented by other modalities.

\subsection{Effectiveness of Transcript}
M$^{2}$SM incorporates video transcripts to bridge videos and texts with appropriate alignments. This is one of the critical factors for 
proper summarization, 
as demonstrated in Table~\ref{tab_ablation}. To further investigate the nature of transcripts, we quantitatively evaluate their 
relationships with the articles and references, as shown in Table~\ref{tab_transcript}. The results demonstrate that video transcripts are distinct from articles with low overlaps, indicating that they are not repeating
of articles but provide extra and useful information.
Table~\ref{tab_transcript} also illustrates that they poorly correlate with references, which suggests that transcripts assist summary generation by capturing the key information of videos, however they are not enough for final summaries.

\subsection{Manual Evaluation}
200 examples with text and video summarization results were selected, and 5 graduate students were volunteered to evaluate them based on informativeness (Inform) and satisfaction (Satis). Each sample was graded on the scale from 1 to 5, where a higher score is better.
We calculate the average score of each evaluation (Table ~\ref{tab_manual}) shows it further demonstrates the proposed method.



\section{Conclusion}

In this work, we have proposed a multi-modal summarization task 
that generates summaries from 
documents and 
the related videos. We have also 
constructed a content-rich video-containing dataset for future study. A comprehensive evaluation has demonstrated the effectiveness of the 
proposed model and 
individual introduced strategies.\par
Our work can be extended 
in some ways. For example, 
acoustic features could be extracted from acoustic signals and incorporated into our model to provide 
additional complementary information, i.e., sentiment and tone. 
In another example, to further advance user satisfaction, it would be worthwhile to explore generation techniques, 
such as generating a small video accompanied with text description as a summary.
\clearpage

\bibliography{reference}

\begin{thebibliography}{46}
\expandafter\ifx\csname natexlab\endcsname\relax\def\natexlab#1{#1}\fi

\bibitem[{Bahdanau et~al.(2014)Bahdanau, Cho, and Bengio}]{bahdanau2014neural}
Dzmitry Bahdanau, Kyunghyun Cho, and Yoshua Bengio. 2014.
\newblock Neural machine translation by jointly learning to align and
  translate.
\newblock In \emph{ICLR}.

\bibitem[{Bian et~al.(2013)Bian, Yang, and Chua}]{bian2013multimedia}
Jingwen Bian, Yang Yang, and Tat-Seng Chua. 2013.
\newblock Multimedia summarization for trending topics in microblogs.
\newblock In \emph{CIKM}, pages 1807--1812.

\bibitem[{Cao et~al.(2018)Cao, Li, Li, and Wei}]{cao2018retrieve}
Ziqiang Cao, Wenjie Li, Sujian Li, and Furu Wei. 2018.
\newblock Retrieve, rerank and rewrite: Soft template based neural
  summarization.
\newblock In \emph{ACL}, pages 152--161.

\bibitem[{Chen and Zhuge(2018)}]{chen2018abstractive}
Jingqiang Chen and Hai Zhuge. 2018.
\newblock Abstractive text-image summarization using multi-modal attentional
  hierarchical rnn.
\newblock In \emph{EMNLP}, pages 4046--4056.

\bibitem[{Cheng and Lapata(2016)}]{cheng2016neural}
Jianpeng Cheng and Mirella Lapata. 2016.
\newblock Neural summarization by extracting sentences and words.
\newblock In \emph{ACL}, pages 484--494.

\bibitem[{De~Avila et~al.(2011)De~Avila, Lopes, da~Luz~Jr, and
  de~Albuquerque~Ara{\'u}jo}]{de2011vsumm}
Sandra Eliza~Fontes De~Avila, Ana Paula~Brand{\~a}o Lopes, Antonio da~Luz~Jr,
  and Arnaldo de~Albuquerque~Ara{\'u}jo. 2011.
\newblock Vsumm: A mechanism designed to produce static video summaries and a
  novel evaluation method.
\newblock \emph{Pattern Recognition Letters}, 32(1):56--68.

\bibitem[{Dong et~al.(2019)Dong, Yang, Wang, Wei, Liu, Wang, Gao, Zhou, and
  Hon}]{dong2019unified}
Li~Dong, Nan Yang, Wenhui Wang, Furu Wei, Xiaodong Liu, Yu~Wang, Jianfeng Gao,
  Ming Zhou, and Hsiao-Wuen Hon. 2019.
\newblock Unified language model pre-training for natural language
  understanding and generation.
\newblock In \emph{NeurIPS}, pages 13042--13054.

\bibitem[{Ekin et~al.(2003)Ekin, Tekalp, and Mehrotra}]{ekin2003automatic}
Ahmet Ekin, A~Murat Tekalp, and Rajiv Mehrotra. 2003.
\newblock Automatic soccer video analysis and summarization.
\newblock \emph{IEEE Transactions on Image Processing}, 12(7):796--807.

\bibitem[{Erol et~al.(2003)Erol, Lee, and Hull}]{erol2003multimodal}
Berna Erol, D-S Lee, and Jonathan Hull. 2003.
\newblock Multimodal summarization of meeting recordings.
\newblock In \emph{ICME}, volume~3.

\bibitem[{Evangelopoulos et~al.(2013)Evangelopoulos, Zlatintsi, Potamianos,
  Maragos, Rapantzikos, Skoumas, and Avrithis}]{evangelopoulos2013multimodal}
Georgios Evangelopoulos, Athanasia Zlatintsi, Alexandros Potamianos, Petros
  Maragos, Konstantinos Rapantzikos, Georgios Skoumas, and Yannis Avrithis.
  2013.
\newblock Multimodal saliency and fusion for movie summarization based on
  aural, visual, and textual attention.
\newblock \emph{IEEE Transactions on Multimedia}, 15(7):1553--1568.

\bibitem[{He et~al.(2016)He, Zhang, Ren, and Sun}]{he2016deep}
Kaiming He, Xiangyu Zhang, Shaoqing Ren, and Jian Sun. 2016.
\newblock Deep residual learning for image recognition.
\newblock In \emph{CVPR}, pages 770--778.

\bibitem[{Hermann et~al.(2015)Hermann, Kocisky, Grefenstette, Espeholt, Kay,
  Suleyman, and Blunsom}]{hermann2015teaching}
Karl~Moritz Hermann, Tomas Kocisky, Edward Grefenstette, Lasse Espeholt, Will
  Kay, Mustafa Suleyman, and Phil Blunsom. 2015.
\newblock Teaching machines to read and comprehend.
\newblock In \emph{NeurIPS}, pages 1693--1701.

\bibitem[{Hori et~al.(2019)Hori, Alamri, Wang, Wichern, Hori, Cherian, Marks,
  Cartillier, Lopes, Das et~al.}]{hori2019end}
Chiori Hori, Huda Alamri, Jue Wang, Gordon Wichern, Takaaki Hori, Anoop
  Cherian, Tim~K Marks, Vincent Cartillier, Raphael~Gontijo Lopes, Abhishek
  Das, et~al. 2019.
\newblock End-to-end audio visual scene-aware dialog using multimodal
  attention-based video features.
\newblock In \emph{ICASSP}, pages 2352--2356.

\bibitem[{Jacob et~al.(2019)Jacob, Ming-Wei, Kenton, and Kristina}]{bert2019}
Devlin Jacob, Chang Ming-Wei, Lee Kenton, and Toutanova Kristina. 2019.
\newblock Bert: Pre-training of deep bidirectional transformers for language
  understanding.
\newblock In \emph{NAACL}, pages 4171--4186.

\bibitem[{Jadhav and Rajan(2018)}]{jadhav2018extractive}
Aishwarya Jadhav and Vaibhav Rajan. 2018.
\newblock Extractive summarization with swap-net: Sentences and words from
  alternating pointer networks.
\newblock In \emph{ACL}, pages 142--151.

\bibitem[{Kim et~al.(2017)Kim, On, Lim, Kim, Ha, and Zhang}]{kim2017hadamard}
Jin-Hwa Kim, Kyoung-Woon On, Woosang Lim, Jeonghee Kim, Jung-Woo Ha, and
  Byoung-Tak Zhang. 2017.
\newblock Hadamard product for low-rank bilinear pooling.
\newblock In \emph{ICLR}.

\bibitem[{Li et~al.(2020)Li, Yuan, Xu, Wu, He, and Zhou}]{li-aspect}
Haoran Li, Peng Yuan, Song Xu, Youzheng Wu, Xiaodong He, and Bowen Zhou. 2020.
\newblock Aspect-aware multimodal summarization for chinese e-commerce
  products.
\newblock In \emph{AAAI}, page in press.

\bibitem[{Li et~al.(2018)Li, Zhu, Liu, Zhang, and
  Zong}]{li2018multi-modal_sent}
Haoran Li, Junnan Zhu, Tianshang Liu, Jiajun Zhang, and Chengqing Zong. 2018.
\newblock Multi-modal sentence summarization with modality attention and image
  filtering.
\newblock In \emph{IJCAI}, pages 4152--4158.

\bibitem[{Li et~al.(2017)Li, Zhu, Ma, Zhang, and Zong}]{li2017multi-modal}
Haoran Li, Junnan Zhu, Cong Ma, Jiajun Zhang, and Chengqing Zong. 2017.
\newblock Multi-modal summarization for asynchronous collection of text, image,
  audio and video.
\newblock In \emph{EMNLP}, pages 1092--1102.

\bibitem[{Li et~al.(2019)Li, Zhang, Ji, and Radke}]{li-etal-2019-keep}
Manling Li, Lingyu Zhang, Heng Ji, and Richard~J. Radke. 2019.
\newblock Keep meeting summaries on topic: Abstractive multi-modal meeting
  summarization.
\newblock In \emph{ACL}, pages 2190--2196.

\bibitem[{Libovick{\`y} et~al.(2018)Libovick{\`y}, Palaskar, Gella, and
  Metze}]{libovicky2018multimodal}
Jindrich Libovick{\`y}, Shruti Palaskar, Spandana Gella, and Florian Metze.
  2018.
\newblock Multimodal abstractive summarization of opendomain videos.
\newblock In \emph{NeurIPS Workshop on ViGIL}.

\bibitem[{Lin(2004)}]{lin2004rouge}
Chin-Yew Lin. 2004.
\newblock Rouge: A package for automatic evaluation of summaries.
\newblock \emph{Text Summarization Branches Out}, 8.

\bibitem[{Liu et~al.(2018)Liu, Li, Xu, and Natarajan}]{liu2018learn}
Kuan Liu, Yanen Li, Ning Xu, and Prem Natarajan. 2018.
\newblock Learn to combine modalities in multimodal deep learning.
\newblock \emph{arXiv preprint arXiv:1805.11730}.

\bibitem[{Liu and Lapata(2019)}]{liu-lapata-2019-text}
Yang Liu and Mirella Lapata. 2019.
\newblock Text summarization with pretrained encoders.
\newblock In \emph{EMNLP}, pages 3730--3740.

\bibitem[{Lu et~al.(2019)Lu, Batra, Parikh, and Lee}]{lu2019vilbert}
Jiasen Lu, Dhruv Batra, Devi Parikh, and Stefan Lee. 2019.
\newblock Vilbert: Pretraining task-agnostic visiolinguistic representations
  for vision-and-language tasks.
\newblock In \emph{NIPS}, pages 13--23.

\bibitem[{Lu and Grauman(2013)}]{lu2013story}
Zheng Lu and Kristen Grauman. 2013.
\newblock Story-driven summarization for egocentric video.
\newblock In \emph{CVPR}, pages 2714--2721.

\bibitem[{Nallapati et~al.(2017)Nallapati, Zhai, and
  Zhou}]{nallapati2017summarunner}
Ramesh Nallapati, Feifei Zhai, and Bowen Zhou. 2017.
\newblock Summarunner: A recurrent neural network based sequence model for
  extractive summarization of documents.
\newblock In \emph{AAAI}, pages 3075--3081.

\bibitem[{Narayan et~al.(2018)Narayan, Cohen, and Lapata}]{narayan2018ranking}
Shashi Narayan, Shay~B Cohen, and Mirella Lapata. 2018.
\newblock Ranking sentences for extractive summarization with reinforcement
  learning.
\newblock In \emph{NAACL}, pages 1747--1759.

\bibitem[{Palaskar et~al.(2019)Palaskar, Libovick{\'y}, Gella, and
  Metze}]{palaskar-etal-2019-multimodal}
Shruti Palaskar, Jind{\v{r}}ich Libovick{\'y}, Spandana Gella, and Florian
  Metze. 2019.
\newblock Multimodal abstractive summarization for how2 videos.
\newblock In \emph{ACL}, pages 6587--6596.

\bibitem[{See et~al.(2017)See, Liu, and Manning}]{see2017get}
Abigail See, Peter~J Liu, and Christopher~D Manning. 2017.
\newblock Get to the point: Summarization with pointer-generator networks.
\newblock In \emph{ACL}, pages 1073--1083.

\bibitem[{Snoek et~al.(2005)Snoek, Worring, and Smeulders}]{snoek2005early}
Cees~GM Snoek, Marcel Worring, and Arnold~WM Smeulders. 2005.
\newblock Early versus late fusion in semantic video analysis.
\newblock In \emph{ACM MM}, pages 399--402.

\bibitem[{Song et~al.(2015)Song, Vallmitjana, Stent, and
  Jaimes}]{song2015tvsum}
Yale Song, Jordi Vallmitjana, Amanda Stent, and Alejandro Jaimes. 2015.
\newblock Tvsum: Summarizing web videos using titles.
\newblock In \emph{CVPR}, pages 5179--5187.

\bibitem[{Sun et~al.(2019)Sun, Myers, Vondrick, Murphy, and
  Schmid}]{sun2019videobert}
Chen Sun, Austin Myers, Carl Vondrick, Kevin Murphy, and Cordelia Schmid. 2019.
\newblock Videobert: A joint model for video and language representation
  learning.
\newblock In \emph{ICCV}, pages 7464--7473.

\bibitem[{Tan et~al.(2017)Tan, Wan, and Xiao}]{tan2017abstractive}
Jiwei Tan, Xiaojun Wan, and Jianguo Xiao. 2017.
\newblock Abstractive document summarization with a graph-based attentional
  neural model.
\newblock In \emph{ACL}, pages 1171--1181.

\bibitem[{Tjondronegoro et~al.(2011)Tjondronegoro, Tao, Sasongko, and
  Lau}]{tjondronegoro2011multi}
Dian Tjondronegoro, Xiaohui Tao, Johannes Sasongko, and Cher~Han Lau. 2011.
\newblock Multi-modal summarization of key events and top players in sports
  tournament videos.
\newblock In \emph{WACV}, pages 471--478.

\bibitem[{Truong and Lauw(2019)}]{vistanet}
Quoc-Tuan Truong and Hady~W Lauw. 2019.
\newblock Vistanet: Visual aspect attention network for multimodal sentiment
  analysis.
\newblock In \emph{AAAI}, pages 305--312.

\bibitem[{UzZaman et~al.(2011)UzZaman, Bigham, and
  Allen}]{UzZaman2011multimodal}
Naushad UzZaman, Jeffrey~P. Bigham, and James~F. Allen. 2011.
\newblock Multimodal summarization of complex sentences.
\newblock In \emph{IUI}, pages 43--52.

\bibitem[{Wei et~al.(2018)Wei, Ni, Yan, Yu, Yang, and Yao}]{wei2018video}
Huawei Wei, Bingbing Ni, Yichao Yan, Huanyu Yu, Xiaokang Yang, and Chen Yao.
  2018.
\newblock Video summarization via semantic attended networks.
\newblock In \emph{AAAI}, pages 216--223.

\bibitem[{Wu and Hu(2018)}]{wu2018learning}
Yuxiang Wu and Baotian Hu. 2018.
\newblock Learning to extract coherent summary via deep reinforcement learning.
\newblock In \emph{AAAI}, pages 5602--5609.

\bibitem[{Xiao et~al.(2020)Xiao, Zhao, Zhang, Yan, and
  Yang}]{xiao2020convolutional}
Shuwen Xiao, Zhou Zhao, Zijian Zhang, Xiaohui Yan, and Min Yang. 2020.
\newblock Convolutional hierarchical attention network for query-focused video
  summarization.
\newblock \emph{arXiv preprint arXiv:2002.03740}.

\bibitem[{Zadeh et~al.(2017)Zadeh, Chen, Poria, Cambria, and
  Morency}]{zadeh-etal-2017-tensor}
Amir Zadeh, Minghai Chen, Soujanya Poria, Erik Cambria, and Louis-Philippe
  Morency. 2017.
\newblock Tensor fusion network for multimodal sentiment analysis.
\newblock In \emph{EMNLP}, pages 1103--1114.

\bibitem[{Zhang et~al.(2019)Zhang, Wei, and Zhou}]{zhang-etal-2019-hibert}
Xingxing Zhang, Furu Wei, and Ming Zhou. 2019.
\newblock {HIBERT}: Document level pre-training of hierarchical bidirectional
  transformers for document summarization.
\newblock In \emph{ACL}, pages 5059--5069.

\bibitem[{Zhou et~al.(2018)Zhou, Qiao, and Xiang}]{zhou2018deep}
Kaiyang Zhou, Yu~Qiao, and Tao Xiang. 2018.
\newblock Deep reinforcement learning for unsupervised video summarization with
  diversity-representativeness reward.
\newblock In \emph{AAAI}, pages 7582--7589.

\bibitem[{Zhou et~al.(2020)Zhou, Palangi, Zhang, Hu, Corso, and
  Gao}]{zhou2019unified}
Luowei Zhou, Hamid Palangi, Lei Zhang, Houdong Hu, Jason~J Corso, and Jianfeng
  Gao. 2020.
\newblock Unified vision-language pre-training for image captioning and vqa.
\newblock In \emph{AAAI}, page in press.

\bibitem[{Zhu et~al.(2018)Zhu, Li, Liu, Zhou, Zhang, and Zong}]{zhu2018msmo}
Junnan Zhu, Haoran Li, Tianshang Liu, Yu~Zhou, Jiajun Zhang, and Chengqing
  Zong. 2018.
\newblock Msmo: Multimodal summarization with multimodal output.
\newblock In \emph{EMNLP}, pages 4154--4164.

\bibitem[{Zhu et~al.(2020)Zhu, Zhou, Zhang, Li, Zong, and Li}]{zhu3multimodal}
Junnan Zhu, Yu~Zhou, Jiajun Zhang, Haoran Li, Chengqing Zong, and Changliang
  Li. 2020.
\newblock Multimodal summarization with guidance of multimodal reference.
\newblock In \emph{AAAI}, page in press.

\end{thebibliography}
\bibliographystyle{acl_natbib}

\end{document}